# Type and Complexity Signals in Multilingual Question Representations

**Robin Kokot** and **Wessel Poelman**
Department of Computer Science,
KU Leuven
`{robin.kokot, wessel.poelman}@kuleuven.be`

## Abstract

This work investigates how a multilingual transformer model represents morphosyntactic properties of questions. We introduce the Question Type and Complexity (QTC) dataset with sentences across seven languages, annotated with type information and complexity metrics including dependency length, tree depth, and lexical density. Our evaluation extends probing methods to regression labels with selectivity controls to quantify gains in generalizability. We compare layer-wise probes on frozen Glot500-m (Imani et al., 2023) representations against subword TF-IDF baselines, and a fine-tuned model. Results show that statistical features classify questions effectively in languages with explicit marking, while neural probes capture fine-grained structural complexity patterns better. We use these results to evaluate when contextual representations outperform statistical baselines and whether parameter updates reduce availability of pre-trained linguistic information.

## 1 Introduction

Multilingual contextual embeddings show promise for accessing fine-grained morphosyntactic properties across hundreds of languages. Probing how transformer models encode certain linguistic properties has practical implications for language typology research, where systematic comparison of structural features often relies on automated analysis. Additionally, evaluations targeting specific linguistic phenomena can test common architectural assumptions about transformer models. Examples include the often discussed layer-wise specialization from syntactic to semantic processing (Tenney et al., 2019a) and the ability of shared embedding spaces to effectively capture cross-linguistic patterns.

Researchers rely on these assumptions in order to describe the internals of the models when testing on benchmarks (Conneau et al., 2020; Şahin et al., 2020), but also when evaluating their general linguistic capabilities outside of specific tasks (Brunato et al., 2020). However, comparisons with appropriate baselines are often left out. Without those, we cannot determine whether observed linguistic capabilities reflect genuine structural processing or are the result of patterns that simpler statistical methods capture equally well.

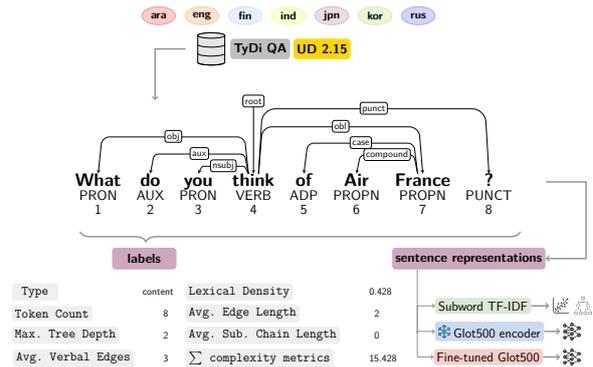

Figure 1: Experimental pipeline from multilingual datasets: TyDi QA (Clark et al., 2020), UD 2.15 (Zeman et al., 2024), through annotation of question types and complexity metrics to extraction of three representation types used for model training.

This presents a challenge when investigating universal sentence-level phenomena where the relationship between surface form and underlying structure varies extensively (Tenney et al., 2019b; Ravishankar et al., 2019). We focus specifically on *interrogative sentences*, which illustrate this variation particularly well. For example, Arabic uses explicit particles like "هل" for polar (yes/no) questions and overt subordinating conjunctions for complex clauses. Alternatively, Japanese relies on contextual cues for question interpretation and implicit hierarchical embedding through case-marking for structural complexity. The differences in how languages encode both categorical distinctions and continuous complexity metrics create a natural setup for evaluating whether contextual rep-resentations capture structural patterns better than surface-level statistical correlations obtained by, for example, TF-IDF features.

We explore this question through controlled comparisons of neural representations with statistical baselines across seven typologically diverse languages. Our framework extends probing methods to continuous linguistic properties, including token count, lexical density, dependency length, tree depth, verbal arity, and subordination patterns. Figure 1 illustrates our method: we start with existing multilingual datasets, process and annotate these for categorical (interrogative types) and continuous labels, and we finally evaluate three representation types (subword TF-IDF features, contextual embeddings, and a fine-tuned model) using our annotated data in Arabic, English, Finnish, Indonesian, Japanese, Korean, and Russian.

We present three key findings:

- Contextual embeddings outperform statistical baselines for question type classification, particularly in languages requiring contextual integration (Japanese, Korean, English, Finnish).
- Regression performance varies significantly across metrics, with distinct layer-wise profiles emerging for different structural properties.
- Fine-tuning compensates for unstable neural encoding patterns but degrades performance on metrics with stable layer-wise representations, revealing a trade-off between adaptation and preservation of pre-trained linguistic knowledge.

These results provide practical guidance for model selection based on typological properties and suggest that frozen representations may be preferable for certain analytical tasks. Additionally, our regression-based probing framework with selectivity controls opens new avenues for investigating continuous linguistic properties in neural representations.

## 2 Related Work

Probing methods assess what linguistic knowledge is encoded in neural representations by training classifiers to predict specific properties in word embeddings (Adi et al., 2017; Conneau et al., 2018). Early work demonstrated that contextualized and static representations encode syntactic information like part-of-speech categories, dependency relations, and word order variation (Köhn, 2015; Shi et al., 2016).

Most probing studies focus on token-level properties, with fewer approaches looking at variation in sentence-level regularities. Şahin et al. (2020); Waldis et al. (2024) introduce comprehensive evaluation frameworks for sentence level probing tasks. These reveal how models encode structural linguistic properties such as morphological case marking, agreement patterns, and syntactic hierarchies, as well as functional properties including semantic roles, discourse relations, and pragmatic features. Question type classification represents a natural extension of this work, as it requires models to integrate both formal markers (interrogative particles, auxiliary inversion) and functional understanding (information-seeking intent, presupposition structure).

Two assumptions motivate current probing approaches. First, the layer-wise specialization hypothesis suggests lower layers encode syntax while higher layers capture semantics (Tenney et al., 2019a). This informs decisions about which layers to probe for different linguistic tasks. Second, multilingual models develop shared embedding spaces that capture cross-linguistic patterns (Conneau et al., 2020), enabling efficient transfer across languages.

Probes mainly target categorical properties through classification tasks (Tenney et al., 2019b; Jawahar et al., 2019). However, Pimentel et al. (2020) argue that complex linguistic phenomena require more sophisticated probing architectures that can approximate a wider range of information content. Regression-based probing is a simple approach that investigates linguistic properties like syntactic complexity, processing difficulty, and structural density. Complexity measures derived through dependency parsing allow us to generate target labels that reveal how models encode syntactic structure along continuous and discrete dimensions. We investigate these to assess how accessible structural features are from learned embeddings.

Determining whether probes capture genuine structural encoding requires appropriate baselines. Hewitt and Liang (2019) introduced selectivity controls comparing performance on real versus shuffled labels to distinguish linguistic encoding from spurious correlations. Most studies, however, evaluate neural representations without statistical baselines, making it difficult to assess whether contextual embeddings offer genuine advantages over frequency-based methods.

| Language   | #     | % Polar | % Content | Avg. Score |
|------------|-------|---------|-----------|------------|
| Arabic     | 1,116 | 48.3    | 51.7      | 0.42       |
| English    | 1,374 | 50.0    | 50.0      | 0.38       |
| Finnish    | 1,368 | 49.9    | 50.1      | 0.34       |
| Indonesian | 1,136 | 48.2    | 51.8      | 0.39       |
| Japanese   | 1,329 | 50.8    | 49.2      | 0.41       |
| Korean     | 921   | 46.9    | 53.1      | 0.39       |
| Russian    | 1,376 | 50.0    | 50.0      | 0.41       |
| **Total**  | **8,620** | **50.6** | **49.4** | **0.39** |

Table 1: QTC dataset statistics by language. # shows total annotated sentences per language. Polar/Content percentages reflect question type distribution. Average Complexity represents normalized composite scores of individual complexity metrics (see Appendix A for details).

Similarly, while probing typically uses frozen representations extracted from a specific layer of the encoder, the relationship between pre-trained knowledge and task-specific optimization remains underexplored. Understanding when end-to-end optimization preserves or degrades pre-trained linguistic knowledge requires direct comparison of frozen probes and fine-tuned models, particularly for structural properties that may be disrupted by task-specific adaptation.

## 3 Dataset

We introduce the Question Type and Complexity (QTC) dataset containing ∼ 9,000 annotated questions across seven languages: Arabic, English, Finnish, Indonesian, Japanese, Korean, and Russian. QTC combines TyDiQA-GoldP training data (Clark et al., 2020) with Universal Dependency treebank test data (Nivre et al., 2020; Zeman et al., 2024) to balance natural language variation with standardized syntactic annotation, with approximately 70% of sentences drawn from TyDiQA and 30% from UD treebanks. The choice of languages was informed by different question formation strategies. Languages using explicit interrogative marking include Arabic with "هل", Finnish with suffix "-ko/-kö", and Russian with particle "ли". Languages using implicit strategies like context or prosody include Japanese, Indonesian, and Korean. Lastly, auxiliary inversion in English can be seen as a mixed strategy.

Categorical and continuous labels were created using parallel annotation pipelines. For question type classification, TyDiQA data already contained human annotations from three independent annotators. We adopted annotations where all three annotators agreed and manually resolved disagreements. UD treebank sentences were annotated for question type using language-specific rule-based systems targeting morphosyntactic patterns: interrogative particles, wh-phrase positioning, and auxiliary structures. We label polar questions as '1' and content questions as '0'.

For complexity metrics, we used UDPipe 2.0 (Straka, 2018) to parse all sentences, then applied the Profiling-UD framework (Brunato et al., 2020) to extract six raw complexity features capturing processing difficulty (see Appendix A for details). We validated complexity metrics through statistical outlier detection and (partial) manual verification of parse quality.[1]

## 4 Probing Tasks

### 4.1 Question Type Classification

Classifying questions as polar (yes-no) or content (wh-) is an interesting test case for comparing neural representations against statistical baselines. As mentioned, languages with explicit marking strategies use dedicated particles or consistent transformations, like English auxiliary inversion (Dryer, 2013a). This makes classes identifiable through surface patterns that frequency features can capture.

Languages with implicit strategies prove challenging because they rely on context and prosody. Japanese polar questions like "Ashita kimasu ka?" [Tomorrow come-polite Q] and content questions "Itsu kimasu ka?" [When come-polite Q] have identical sentence-final particles, differing only in the presence of wh-words that often appear in non-initial positions (Dryer, 2013b). This variation allows us to test when contextual embeddings provide genuine advantages over frequency-based approaches for capturing structural patterns that go beyond readily available surface cues.

### 4.2 Linguistic Complexity Prediction

In addition to question type classification, we also use *continuous* labels and predict complexity scores derived from morphosyntactic properties. This operationalizes the idea that structural density increases processing difficulty (Hawkins, 2007). We formulate this as a regression task, targeting six normalized complexity metrics: token count, lexical density, average dependency length, maximum tree depth, verbal arity, and subordinate chain length. This tests whether different representations capture quantitative aspects of linguistic structure. We also evaluate performance on a combined complexity score calculated as the normalized sum of all six individual metrics, providing an abstract measure of structural density.

---
[1]The QTC dataset and code are available at hf.co/rokokot/question-type-and-complexity and github.com/rokokot/qtype-eval.

Statistical models can effectively capture surface-level complexity indicators. Token count correlates with question length from simple "Who left?" to complex "What did the committee decide about the proposal?", while subordination patterns manifest through explicit conjunctions that TF-IDF features can detect. However, hierarchical syntactic properties present greater challenges. A question like "Who ate the cake that Alice brought?" shares the same interrogative markers as the simple example, but involves multiple dependency levels and clauses that increase syntactic complexity.

Unlike categorical properties typically studied in probing, continuous dimensions allow us to isolate aspects of linguistic structure most effectively captured by different representation approaches. This allows us to test competing hypotheses about how neural and statistical models encode structural information. If contextual representations truly capture abstract syntactic hierarchies, they should outperform frequency-based methods on metrics like tree depth and subordination complexity, which require understanding of long-distance dependencies and recursive structures. Conversely, if neural advantages primarily reflect sophisticated pattern matching, we expect statistical baselines to remain competitive across all complexity dimensions.

### 4.3 Experimental Setup

Our setup addresses the core methodological challenge of distinguishing genuine linguistic encoding from pattern memorization when comparing neural and statistical approaches. Following Hewitt and Liang (2019), we create three shuffled-label control variants per task that preserve label distributions while destroying text-label relationships. We define selectivity as normalized performance differences:

$$S_{cls} = \frac{\text{acc}_{\text{real}} - \text{acc}_{\text{control}}}{\text{acc}_{\text{control}}}$$
$$S_{reg} = \frac{\text{mse}_{\text{control}} - \text{mse}_{\text{real}}}{\text{mse}_{\text{control}}} \quad (1)$$

with (acc)uracy for the classification task and mean squared error (mse) for regression task.

This approach enables direct comparison of representational quality. Selectivity measures how much better a model performs when linguistic structure is present versus absent. Higher values (e.g., $> 0.5$) mean the model exploits "genuine" linguistic patterns, while low selectivity suggests the model performs similarly regardless of whether input-label relationships are meaningful or random. Strong selectivity shows when models capture information rather than surface correlations.

## 5 Experiments

The experiments were carried out on Glot500-m (Imani et al., 2023), a multilingual encoder-only transformer. Glot500-m was created by extending the XLM-R-base architecture (Conneau et al., 2020) using continued pre-training on a custom multilingual corpus and expanding the vocabulary from 250K to 401K tokens to cover 511 languages, including all seven languages in our dataset.

### 5.1 Subword TF-IDF Baselines

First, we establish baselines using linear and nonlinear predictors trained on TF-IDF features and corresponding sentence labels. We use the Glot-500-m tokenizer to generate TF-IDF representations for a fair comparison.

We establish baselines using linear models (logistic regression for classification, ridge regression for complexity prediction) and XGBoost (Chen and Guestrin, 2016) for nonlinear feature interactions. XGBoost provides an upper bound for statistical baseline performance while maintaining interpretability through feature importance scores. Dummy baselines using majority class and mean value prediction set floor performance.

### 5.2 Probes on Frozen Representations

We extract sentence-level embeddings from each of the 12 layers of the frozen encoder using mean pooling across token representations, resulting in a fixed-size 768-dimensional vector for each sentence. For every sentence embedding at every layer we train neural probes to predict the target label. This allows us to track where different kinds of linguistic information are most accessible to the probe.

We designed our probe architectures to capture complex patterns while maintaining training efficiency. Classification probes use two-layer MLPs with 384 hidden units optimized using binary cross-entropy loss. Regression probes use three-layer MLPs with 128 hidden units and minimize the mean squared error loss. All probes are trained separately for each layer and task combination using 70/15/15 train/validation/test splits. While expressive enough to capture complex patterns, this setup ensures that performance differences reflect representational properties rather than probe capacity (Pimentel et al., 2020; Waldis et al., 2024).

| Language | TF-IDF Linear | $\overline{S}$ | TF-IDF XGBoost | $\overline{S}$ | Glot500 Best Probe | $\overline{S}$ | Layer | Glot500 Fine-tuned |
|---|---|---|---|---|---|---|---|---|
| ara Arabic | 90.9 | **0.92** | **97.4** | 0.83 | 85.7 | 0.20 | 2 | 74.1 |
| eng English | 83.6 | 0.55 | 80.9 | 0.56 | **97.3** | **0.95** | 5 | 91.8 |
| fin Finnish | 84.5 | 0.85 | 87.2 | 0.91 | **94.5** | **0.89** | 5 | 92.3 |
| ind Indonesian | 67.3 | 0.41 | 65.5 | 0.23 | **80.9** | **0.62** | 6 | 73.6 |
| jpn Japanese | 64.1 | 0.25 | 64.1 | 0.28 | 82.6 | **1.07** | 10 | **88.0** |
| kor Korean | 66.3 | 0.43 | 73.6 | 0.48 | 76.4 | **0.53** | 9 | **91.1** |
| rus Russian | 86.4 | 0.85 | 77.2 | 0.5 | **97.3** | **0.95** | 11 | 96.4 |

Table 2: Question type classification accuracy (%) and mean selectivity ($\overline{S}$) across approaches. Bold values indicate the highest accuracy and selectivity scores achieved for each language. Layer denotes the index of the encoder layer at which the probe achieved highest accuracy.

### 5.3 Fine-tuned Model

To determine whether parameter updates preserve pre-trained linguistic information, we train the complete Glot500 model end-to-end on each task. The fine-tuned model uses identical task-specific heads as our probes but allows model updates (i.e., not frozen).

We employ two-layer MLPs with binary cross-entropy loss for classification and three-layer heads with MSE loss for regression.

This configuration enables direct comparison with frozen probes. If fine-tuning enhances linguistic representations, the updated model should consistently outperform probes across all metrics. Conversely, degraded performance indicates that task-specific optimization disrupts structural knowledge encoded during pre-training.

We only report main task performance metrics for fine-tuned models because selectivity controls are less meaningful when the entire network adapts to the specific label distribution, potentially reflecting task-specific overfitting.

## 6 Results

Our statistical baselines employ logistic regression for classification and ridge regression for complexity prediction, with XGBoost capturing nonlinear feature interactions.

Results across the two tasks reveal trade-offs in the ability of our models to capture different kinds of linguistic information. For question type classification, neural probes consistently perform the best, with the majority of highest accuracy and selectivity scores. Regression results show more variety, with different representation types leading on different complexity metrics.

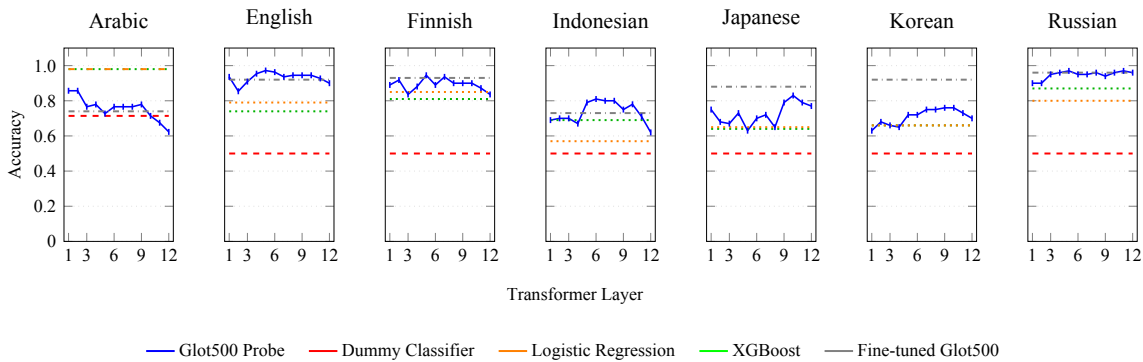

Figure 2: Question type classification across languages and methods. Probing results per layer of Glot500-m.

| Submetric | TF-IDF Ridge | $\overline{S}$ | TF-IDF XGBoost | $\overline{S}$ | Glot500 Best Probe | $\overline{S}$ | Layer | Glot500 Fine-tuned |
|---|---|---|---|---|---|---|---|---|
| Token Count | 0.042 | 0.46 ara | 0.032 | 0.60 ara | **0.004** | 0.68 rus | 5 | 0.006 rus |
| Max. Tree Depth | 0.013 | **0.60** eng | 0.013 | 0.58 eng | **0.002** | 0.57 kor | 2 | 0.017 rus |
| Avg. Dependency Length | 0.007 | 0.36 eng | 0.007 | **0.73** eng | 0.013 | 0.29 rus | 4 | **0.002** fin |
| Avg. Subordinate Chain Length | **0.015** | 0.52 eng | 0.055 | 0.29 ind | 0.053 | 0.47 jpn | 6 | 0.019 eng |
| Avg. Verbal Edges | 0.042 | 0.35 eng | 0.066 | 0.32 fin | 0.070 | 0.40 jpn | 6 | **0.030** ara |
| Lexical Density | 0.033 | 0.48 ind | 0.082 | 0.27 kor | 0.036 | 0.21 ind | 3 | **0.023** eng |
| Combined Complexity | 0.032 | 0.48 rus | 0.017 | 0.60 eng | **0.016** | 0.78 jpn | 4 | 0.020 eng |

Table 3: Complexity submetric regression errors (mse) and mean selectivity ($\overline{S}$) across approaches. Language codes are shown next to every $\overline{S}$ value to indicate the corresponding language.

## 6.1 Surface Markers and Contextual Classification Cues

Table 2 shows the classification accuracy and selectivity scores across all languages and predictors. Probes achieve the highest accuracy in four out of seven languages and the best selectivity scores in six. Arabic is the exception with XGBoost reaching 97.4% accuracy (0.83 selectivity) compared to 85.7% accuracy (0.20 selectivity) with the best performing probe. Linear models perform similarly well (90.9% accuracy, 0.92 selectivity).

Figure 2 tracks how probes perform when trained on representations from different encoder layers, compared to baseline predictors and the fine-tuned model. English, Finnish and Russian show similar trends, with both probes and fine-tuning achieving accuracies > 90%, although at different depths (layer 5 for English and Finnish, layer 11 for Russian).

Indonesian probes perform poorly until layer 5, after which they consistently exceed all baseline methods, dipping only at the final layer. Japanese and Korean show oscillating scores across layers, with fine-tuning achieving notably higher accuracy.

The benefits of contextual representations are clearest in English, Japanese, and Korean, where the performance gap between statistical baselines and Glot500-m probes/fine-tuning ranges from 10 to 20 percentage points increases. Finnish shows a more moderate contextual advantage of less than 10 percentage points, while Arabic, Indonesian, and Russian exhibit much smaller gaps between representation types.

## 6.2 Continuous Complexity Probing

Table 3 presents regression errors across six complexity sub-metrics plus the combined complexity score, limited to results for languages that achieved the best performance on each metric.

Glot500-m probes achieve the lowest error rates on three metrics: token count (0.004 MSE, 0.68 selectivity), tree depth (0.002 MSE, 0.57 selectivity), and combined complexity (0.016 MSE, 0.78 selectivity). Fine-tuning leads on three others: dependency length (0.002 MSE), verbal edges (0.030 MSE), and lexical density (0.023 MSE). Ridge regression achieves the best performance on subordinate chain length (0.015 MSE, 0.52 selectivity).

In terms of selectivity, statistical approaches are surprisingly competitive, with TF-IDF methods achieving the highest selectivity on four out of seven metrics. This contrasts with classification results where probes consistently outperformed our baselines.

Layer-wise regression patterns come in three distinct profiles. Most combinations show flat performance curves where all approaches converge around similar values, with the difference between highest and lowest error remaining below 0.01. Cases with moderate layer-to-layer variation (error differences between 0.01 and 0.03) suggest partial encoding of relevant information across the model's depth. More pronounced oscillations, where error differences exceed 0.03, are usually coupled with low probe performance and point to failures of the contextual embeddings to encode the targeted information.

Fine-tuning achieves the lowest error rates on three metrics: dependency length, verbal edges, lexical density. These advantages appear concentrated on metrics that show relatively flat layer-wise profiles, suggesting that the linguistic properties may be better captured through end-to-end optimization rather than frozen representations. Conversely, metrics where probes excel (token count, tree depth, combined complexity) tend to show more pronounced layer preferences, with fine-tuning performing relatively poorly.

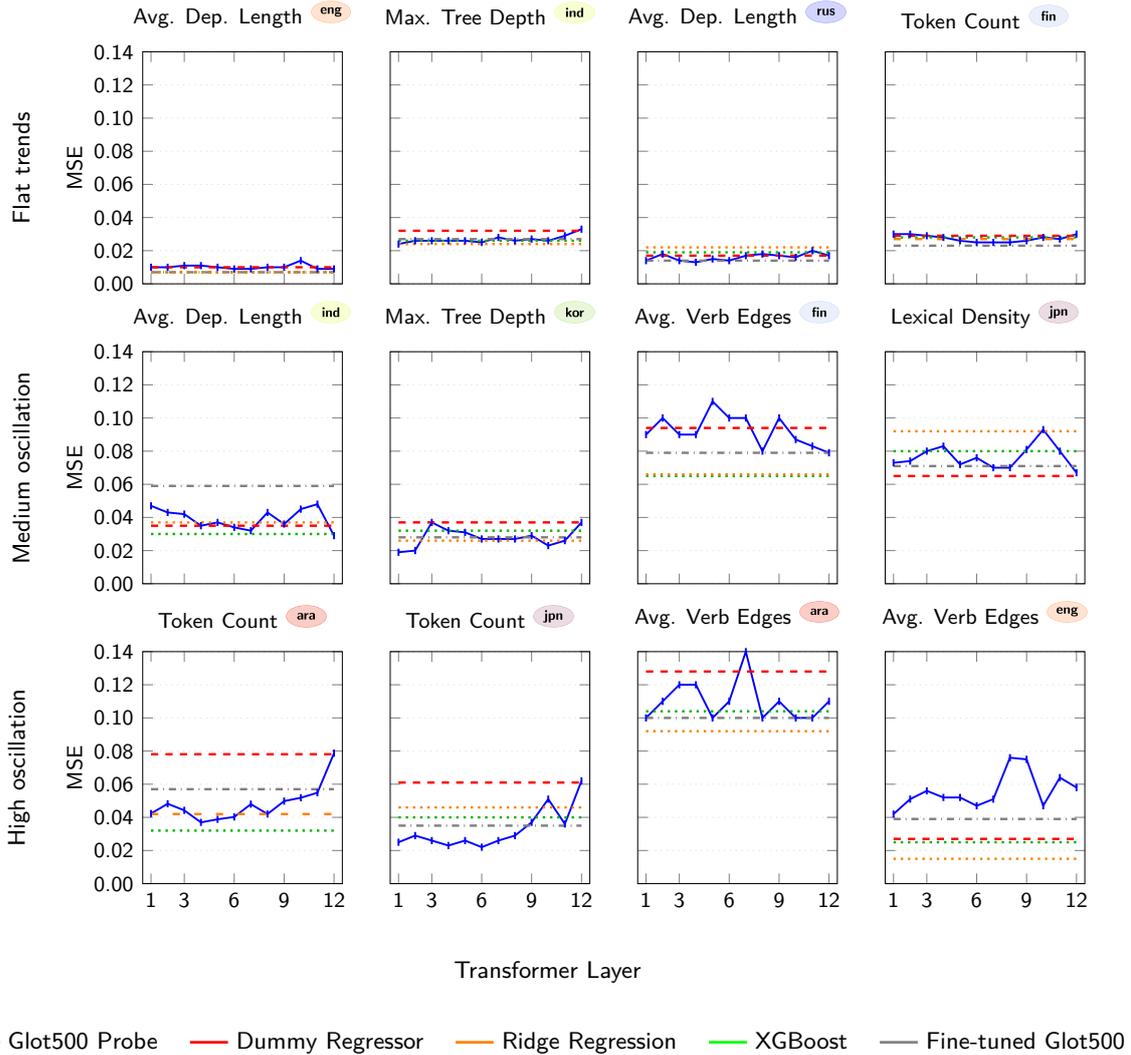

Figure 3: Layer-wise regression trends show three distinct encoding profiles.

## 7 Discussion

### 7.1 Classification Performance Analysis

Classification results show that neural approaches generally perform better than statistical baselines. Arabic is the only exception, possibly reflecting the widespread and unambiguous use of question particles. However, explicit marking does not uniformly favor TF-IDF features. For example, Russian uses particles which follow similar transformations, yet benefit more from Glot500 embeddings. This confirms that the nature of marking strategy matters more than its presence or absence.

English and Finnish further confirm this pattern. English auxiliary inversion ("Is it raining?" / "It is raining") and Finnish suffixes ("-ko/-kö") represent complex morphosyntactic transformations rather than simple particles, yet both show strong neural advantages (97.3% and 94.5% respectively) over statistical baselines. This suggests that transformational marking strategies require contextual processing to identify the relevant structural changes, even when the markers themselves are explicit.

Selectivity scores reveal that these representation types capture distinct aspects of question formation. Statistical methods excel when surface distributions provide reliable cues, while neural representations become necessary when question identification requires integration of distributed contextual information that goes beyond simple frequency patterns.

Differences in processing stability across languages can be seen in Figure 2. These patterns appear related to how question type information is distributed through the transformer architecture. Arabic exhibits the highest variability with 0.25 accuracy difference between layers following its overall lower neural performance. Russian has the opposite tendency with minimal variation ($< 0.1$) while maintaining consistently high performance. English, Korean, and Finnish show moderate variability (0.1), while Indonesian and Japanese display higher fluctuations ($> 0.2$) that correspond with the oscillations visible in their layer-wise profiles.

### 7.2 Regression Profiles

Despite Glot500 representations achieving consistently low errors in complexity regression tasks, TF-IDF approaches show higher selectivity scores on tree depth, dependency length, subordinate chain length, and lexical density metrics. Table 3 shows the variety in the ability of transformers to encode different morphosyntactic properties.

Regression probes have a clear advantage on token count and combined complexity, but they do not consistently outperform our baselines on most of the metrics. Selectivity scores reveal that statistical methods lead on four of seven complexity metrics, demonstrating that frequency features can distinguish meaningful structural patterns from spurious correlations more reliably than contextual representations.

Figure 3 shows a selection of layer-wise error trends highlighting the three most common performance profiles. Flat probe trends were observed most often, meaning that certain structural properties are tied to surface features and that contextual processing rarely provides additional benefit. High probe oscillations and poor performance are more interesting. They imply that rather than building increasingly sophisticated representations of structural complexity, the transformer may be losing and regaining access to relevant information as we move through successive layers.

The differences between fine-tuning and frozen probes point towards a trade-off between the two neural approaches. Fine-tuning performs well almost exclusively on metrics characterized by high oscillations and unstable layer-wise trends, suggesting that parameter updates may compensate for inconsistencies. On the other hand, low performance on metrics with flat profiles demonstrates that task-specific training may prevent access to or even destroy pre-trained information. In other words, when structural information is clearly encoded at specific layers, the parameter updates required for task optimization appear to interfere with these patterns.

However, the success of fine-tuning on predicting dependency length, verbal edges, and lexical density suggests that some properties are not readily available in frozen transformer representations, requiring parameter updates to achieve reliable performance on these metrics.

## 8 Conclusion

We investigated how multilingual transformers encode question patterns by comparing contextual embeddings against statistical baselines across seven typologically diverse languages. Glot500 probes show advantages in question type classification, particularly for languages requiring contextual integration (Japanese, Korean, English, Finnish), while Arabic's unambiguous particles favor statistical methods. For complexity regression, statistical baselines show better selectivity on most individual metrics, though neural methods excel at token count and verbal arity.

Different complexity metrics exhibit distinct layer-wise encoding patterns. Fine-tuning compensates for unstable neural encoding (high oscillations) but struggles on metrics with otherwise stable layer-wise representations, suggesting task-specific optimization can disrupt pre-trained knowledge.

Our QTC dataset and regression-based probing setup using selectivity controls provide tools for investigating continuous linguistic properties. We find that understanding when and why neural models capture linguistic structure requires careful comparison with principled baselines. Future work should examine applications to other architectures, investigate why certain complexity metrics resist neural encoding, and develop training procedures that preserve linguistic information while improving task performance.

## Limitations

This study is limited to seven languages for which high-quality treebanks and interrogative sentence data were available. Our dataset focuses exclusively on questions, so the findings do not generalize to other clause types. While we carefully selected the languages to cover different interrogative patterns, we do not cover all typological variation between target languages. Complexity metrics are computed from automatic dependency parses, which can introduce parser-specific biases and reduce comparability. However, the cross-linguistic consistency of our findings suggests that genuine structural differences emerge despite potential parsing noise.

**Ethics Statement**

All source texts used in the dataset were collected from materials in the public domain or under licenses allowing redistribution for research purposes. No personally identifiable information is included. Although our analysis is diagnostic and does not inform system deployment, findings on cross-linguistic performance could be misused to justify reduced attention to underrepresented languages.

**Acknowledgements**

This paper is based on the first author's Master's thesis completed at KU Leuven (Kokot, 2025). We thank Miryam de Lhoneux for guidance throughout this research, and Kushal Jayesh Tatariya and Vincent Vandeghinste for their valuable feedback on the thesis work. We also thank the anonymous reviewers for constructive comments.

WP is funded by a KU Leuven Bijzonder Onderzoeksfonds C1 project with reference C14/23/096. The resources and services used in this work were provided by the VSC (Flemish Supercomputer Center), funded by the Research Foundation - Flanders (FWO) and the Flemish Government.

## A Metric Definitions

**Token Count** is a straightforward way to measure sentence complexity. It refers to the number of processed segments in a sentence, |T|. In languages like English, tokens are words and punctuation marks. In Japanese or Korean, which do not use spaces between words, tokens are aligned with grammatical morphemes rather than orthographic words. Generally, the more tokens a sentence has, the more likely it is to require greater processing efforts (Iavarone et al., 2021).

**Lexical density** is the ratio of content words (nouns, verbs, adjectives, adverbs) to the total number of tokens excluding punctuation. This metric captures the information density of a sentence and often serves as evidence of register difficulty due to its variation across domains.

$$\text{LD} = \frac{|\text{content words}|}{|\text{T}| - |\text{punct}|} = \frac{3}{7} = 0.428 \quad (2)$$

**Average Dependency Length** is the linear distance between words and their syntactic heads, across all dependency links in a sentence. This measure directly reflects cognitive processing load, as longer dependencies require holding more information during processing. Futrell et al. (2015) provide compelling evidence in 37 languages, showing that all human languages maintain shorter dependency lengths than would occur by random chance.

$$\text{ADL} = \frac{1}{N-1} \sum_{\text{token}(1)}^{N-1} |\text{dep}(i) - \text{head}(i)| = \frac{12}{6} = 2 \quad (3)$$

Where $N$ is the number of tokens (i.e., words) without any punctuation and excluding the root of the sentence ($N-1$).

**Maximum Tree Depth** measures the longest path from root to leaf in a dependency structure, revealing how deeply embedded linguistic elements

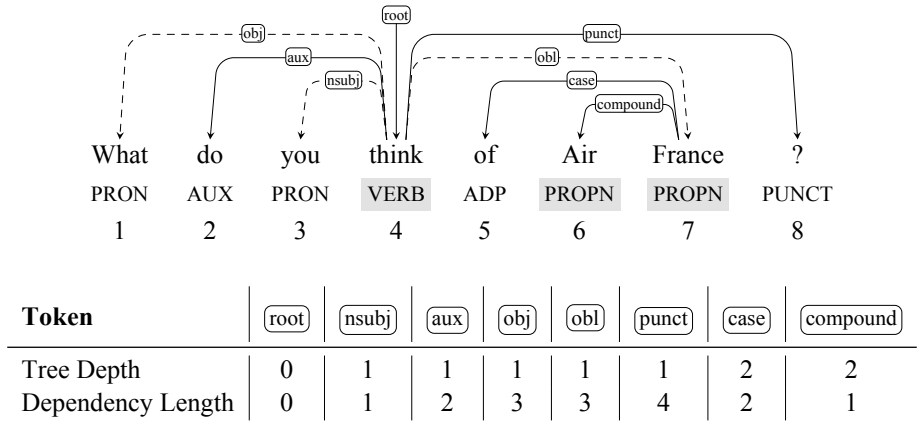

Figure 4: Dependency parse diagram showing grammatical relations, token indices (1-8) and POS tags, with dashed arcs highlighting verbal edges, function words in gray, and a table listing dependency distance and tree depth.

are within a sentence. For token index $i$ in sentence S, Max Depth is defined as:

$$\text{MTD} = \max_{\text{token}(i) \in S} \text{Depth}(i) = 2 \quad (4)$$

**Average Verbal Edges**, sometimes called verbal arity, is a measure of direct dependents (arguments and modifiers) attached to each verb in a sentence. In a dependency structure, these correspond to edges from a verb to its governing words, such as objects, subjects, or adjuncts, but excluding punctuation and auxiliary verbs (Brunato et al., 2020).

$$\overline{\text{ve}} = \frac{1}{|\text{verbs}|} \sum_{v \in \text{verbs}} \text{dependent}(v) = 3 \quad (5)$$

**Average Subordinate Chain Length** is calculated as the ratio of the combined length of all subordinate clauses and the total number of clauses in a sentence. It reflects the level of propositional embedding and recursion. Although the dependency structure in Figure 4 contains no subordination, it remains crucial for capturing the clausal hierarchy of nested sentences.

$$\text{ASC} = \frac{\text{sum of sub. chain lengths}}{\text{number of sub. chains}} = 0 \quad (6)$$

## B  Experimental Details

All experiments were conducted using NVIDIA A100 80GB GPUs. Each probing experiment took approximately 5-10 minutes of training time, while fine-tuning experiments took one hour per language-task combination. The complete experimental suite (including baselines) involved over 3300 individual runs across 7 languages, 12 transformer layers, and multiple tasks with control conditions. This includes experiments with linear algorithms, ensembles, and the fine-tuned model.

Training was carried out with batch sizes of 16, gradient accumulation steps of 2-4, and automatic mixed precision. All experiments used fixed random seeds to ensure reproducibility. The total computational cost was approximately 80 GPU hours on A100 hardware.

For fine-tuning experiments, learning rates were set to 1e-5 for the encoder and 1e-3 for the task head, with early stopping based on validation performance monitored over a patience window of 5 epochs. Probe training used Adam optimizer with learning rate 1e-3 with early stopping when validation loss plateaued.

## C  Additional Results

This appendix provides additional per-language and per-metric performance data to supplement the main analysis. Table 4, Table 5, and Table 6 present detailed results for all seven languages across both classification and regression tasks, including selectivity scores for baseline methods and layer-specific performance indicators for optimal and weakest probe configurations. Figure 5 shows layer-wise error curves for all complexity metrics across languages.

|  |  | Question Type Classification | | | | Combined Complexity Regression | | | |
| --- | --- | --- | --- | --- | --- | --- | --- | --- | --- |
|  |  | Acc. | $\overline{control}$ | $\Delta Acc.$ | $\widehat{S}$ | MSE | $\overline{control}$ | $\Delta MSE$ | $\widehat{S}$ |
| Dummy baseline | ar | 71.4 | 71.4 | 0 | 0 | 0.059 | 0.059 | 0 | 0 |
|  | en | 50.0 | 50.0 | 0 | 0 | 0.038 | 0.039 | 0 | 0 |
|  | fi | 50.0 | 50.0 | 0 | 0 | 0.040 | 0.040 | 0 | 0 |
|  | id | 50.0 | 50.0 | 0 | 0 | 0.040 | 0.040 | 0 | 0 |
|  | ja | 50.0 | 50.0 | 0 | 0 | 0.063 | 0.063 | 0 | 0 |
|  | ko | 50.0 | 50.0 | 0 | 0 | 0.036 | 0.036 | 0 | 0 |
|  | ru | 50.0 | 50.0 | 0 | 0 | 0.059 | 0.059 | 0 | 0 |
| Linear predictors | ar | 90.9 | 47.2 | 43.7 | 0.92 | 0.045 | 0.064 | 0.019 | 0.29 |
|  | en | 83.6 | 53.9 | 29.7 | 0.55 | 0.023 | 0.043 | 0.020 | 0.46 |
|  | fi | 84.5 | 46.1 | 0.83 | 0.85 | 0.037 | 0.043 | 0.006 | 0.14 |
|  | id | 67.3 | 47.5 | 19.8 | 0.41 | 0.028 | 0.046 | 0.018 | 0.39 |
|  | ja | 64.1 | 51.1 | 13.0 | 0.25 | 0.039 | 0.064 | 0.015 | 0.23 |
|  | ko | 66.4 | 46.3 | 20.1 | 0.43 | 0.023 | 0.039 | 0.016 | 0.41 |
|  | ru | 86.4 | 46.7 | 0.85 | 0.85 | 0.032 | 0.069 | 0.037 | 0.53 |
| Gradient boosting | ar | 97.4 | 53.2 | 44.4 | 0.83 | 0.034 | 0.063 | 0.029 | 0.46 |
|  | en | 80.9 | 51.8 | 29.1 | 0.56 | 0.018 | 0.043 | 0.025 | 0.58 |
|  | fi | 87.2 | 45.7 | 41.5 | 0.91 | 0.032 | 0.042 | 0.01 | 0.24 |
|  | id | 65.5 | 53.3 | 12.2 | 0.23 | 0.026 | 0.043 | 0.017 | 0.39 |
|  | ja | 64.1 | 50.0 | 14.1 | 0.28 | 0.037 | 0.061 | 0.024 | 0.39 |
|  | ko | 73.6 | 49.7 | 23.9 | 0.48 | 0.031 | 0.038 | 0.007 | 0.18 |
|  | ru | 77.2 | 48.8 | 28.4 | 0.51 | 0.039 | 0.061 | 0.022 | 0.36 |
| Optimal Probe | ar | 85.7 (2) | 71.4 | 14.3 | 0.20 | 0.030 (4) | 0.067 | 0.037 | 0.55 |
|  | en | 97.3 (5) | 50.0 | 47.3 | 0.95 | 0.017 (1) | 0.048 | 0.031 | 0.64 |
|  | fi | 94.5 (5) | 50.0 | 44.5 | 0.89 | 0.025 (1) | 0.050 | 0.025 | 0.50 |
|  | id | 80.9 (6) | 50.0 | 30.9 | 0.62 | 0.024 (4) | 0.047 | 0.023 | 0.49 |
|  | ja | 82.6 (10) | 39.8 | 42.8 | 1.07 | 0.016 (4) | 0.073 | 0.057 | 0.78 |
|  | ko | 76.4 (9) | 50.0 | 26.4 | 0.53 | 0.043 (3) | 0.093 | 0.050 | 0.53 |
|  | ru | 97.3 (11) | 50.0 | 47.3 | 0.95 | 0.039 (6) | 0.069 | 0.030 | 0.43 |
| Weakest Probe | ar | 62.5 (12) | 71.4 | 8.9 | 0.12 | 0.057 (12) | 0.061 | 0.004 | 0.07 |
|  | en | 89.4 (2) | 50.0 | 39.4 | 0.79 | 0.043 (12) | 0.045 | 0.002 | 0.04 |
|  | fi | 83.6 (3) | 50.0 | 33.6 | 0.67 | 0.042 (12) | 0.043 | 0.001 | 0.03 |
|  | id | 62.7 (12) | 50.0 | 12.7 | 0.25 | 0.042 (12) | 0.046 | 0.004 | 0.09 |
|  | ja | 63.0 (5) | 40.2 | 22.8 | 0.57 | 0.058 (12) | 0.061 | 0.003 | 0.05 |
|  | ko | 63.6 (1) | 50.0 | 13.6 | 0.27 | 0.041 (2) | 0.053 | 0.012 | 0.22 |
|  | ru | 90.0 (2) | 50.0 | 40.0 | 0.80 | 0.067 (10) | 0.075 | 0.008 | 0.11 |
| Fine-tuned Glot500 | ar | 74.1 | - | - | - | 0.042 | - | - | - |
|  | en | 91.8 | - | - | - | 0.020 | - | - | - |
|  | fi | 92.3 | - | - | - | 0.030 | - | - | - |
|  | id | 73.6 | - | - | - | 0.030 | - | - | - |
|  | ja | 88.0 | - | - | - | 0.029 | - | - | - |
|  | ko | 91.1 | - | - | - | 0.031 | - | - | - |
|  | ru | 96.4 | - | - | - | 0.045 | - | - | - |

Table 4: Performance metrics for question type classification (accuracy) and combined complexity regression (MSE) tasks across seven languages.

|  |  | Dependency Length |  | Max. Tree Depth |  | Sub. Chain Length |  |
|---|---|---|---|---|---|---|---|
|  |  | $MSE$ | $\widehat{S}$ | $MSE$ | $\widehat{S}$ | $MSE$ | $\widehat{S}$ |
| Dummy baseline | ar | 0.065 | 0 | 0.052 | 0 | 0.077 | 0 |
|  | en | 0.009 | 0 | 0.029 | 0 | 0.027 | 0 |
|  | fi | 0.026 | 0 | 0.031 | 0 | 0.054 | 0 |
|  | id | 0.036 | 0 | 0.033 | 0 | 0.069 | 0 |
|  | ja | 0.108 | 0 | 0.083 | 0 | 0.092 | 0 |
|  | ko | 0.027 | 0 | 0.037 | 0 | 0.054 | 0 |
|  | ru | 0.017 | 0 | 0.025 | 0 | 0.054 | 0 |
| Linear predictors | ar | 0.045 | 0.29 | 0.028 | 0.48 | 0.054 | 0.34 |
|  | en | 0.007 | 0.3 | 0.013 | 0.60 | 0.015 | 0.51 |
|  | fi | 0.024 | 0.17 | 0.036 | 0 | 0.041 | 0.27 |
|  | id | 0.037 | 0.01 | 0.024 | 0.33 | 0.053 | 0.30 |
|  | ja | 0.09 | 0.18 | 0.065 | 0.22 | 0.076 | 0.25 |
|  | ko | 0.022 | 0.23 | 0.025 | 0.36 | 0.044 | 0.31 |
|  | ru | 0.022 | 0 | 0.019 | 0.29 | 0.035 | 0.42 |
| Gradient boosting | ar | 0.057 | 0.15 | 0.028 | 0.44 | 0.059 | 0.24 |
|  | en | 0.007 | 0.3 | 0.014 | 0.55 | 0.025 | 0.10 |
|  | fi | 0.027 | 0.03 | 0.022 | 0.29 | 0.051 | 0.11 |
|  | id | 0.04 | 0 | 0.026 | 0.21 | 0.055 | 0.29 |
|  | ja | 0.105 | 0.04 | 0.063 | 0.24 | 0.081 | 0.14 |
|  | ko | 0.029 | 0.05 | 0.032 | 0.18 | 0.063 | 0.01 |
|  | ru | 0.019 | 0 | 0.017 | 0.35 | 0.044 | 0.22 |
| Optimal Probe | ar | 0.045 (6) | 0.42 | 0.028 (6) | 0.48 | 0.069 (6) | 0.12 |
|  | en | 0.090 (6) | 0.23 | 0.016 (8) | 0.46 | 0.022 (6) | 0.25 |
|  | fi | 0.025 (8) | 0.18 | 0.016 (7) | 0.52 | 0.047 (12) | -0.01 |
|  | id | 0.030 (12) | 0 | 0.024 (1) | 0.34 | 0.049 (1) | 0.46 |
|  | ja | 0.087 (1) | 0.10 | 0.072 (9) | 0.10 | 0.053 (6) | 0.47 |
|  | ko | 0.023 (8) | 0.18 | 0.020 (2) | 0.57 | 0.047 (5) | 0.26 |
|  | ru | 0.013 (4) | 0.29 | 0.016 (6) | 0.41 | 0.049 (5) | 0.14 |
| Weakest Probe | ar | 0.073 (12) | -0.13 | 0.053 (12) | 0.05 | 0.080 (12) | -0.04 |
|  | en | 0.015 (10) | -0.29 | 0.029 (12) | 0.06 | 0.031 (3) | 0.06 |
|  | fi | 0.030 (2) | 0.07 | 0.037 (12) | -0.05 | 0.073 (10) | 0.08 |
|  | id | 0.049 (11) | 0.16 | 0.033 (12) | 0.05 | 0.081 (11) | 0.06 |
|  | ja | 0.122 (3) | 0.06 | 0.103 (10) | -0.10 | 0.094 (12) | -0.03 |
|  | ko | 0.042 (3) | 0.18 | 0.037 (12) | 0.02 | 0.070 (7) | -0.05 |
|  | ru | 0.020 (11) | 0.07 | 0.028 (12) | 0 | 0.058 (6) | 0.03 |
| Fine-tuned Glot500 | ar | 0.056 | - | 0.038 | - | 0.069 | - |
|  | en | 0.008 | - | 0.022 | - | 0.019 | - |
|  | fi | 0.002 | - | 0.023 | - | 0.045 | - |
|  | id | 0.031 | - | 0.028 | - | 0.043 | - |
|  | ja | 0.105 | - | 0.068 | - | 0.061 | - |
|  | ko | 0.025 | - | 0.029 | - | 0.046 | - |
|  | ru | 0.015 | - | 0.017 | - | 0.046 | - |

Table 5: Performance metrics for linguistic complexity sub-metric regression tasks across seven languages (Part 1: Dependency Length, Tree Depth, Subordinate Chain Length).

|  |  | Verbal Edges | | Lexical Density | | N Tokens | |
|---|---|---|---|---|---|---|---|
|  |  | $MSE$ | $\widehat{S}$ | $MSE$ | $\widehat{S}$ | $MSE$ | $\widehat{S}$ |
| Dummy baseline | ar | 0.060 | 0 | 0.067 | 0 | 0.078 | 0 |
|  | en | 0.060 | 0 | 0.028 | 0 | 0.029 | 0 |
|  | fi | 0.060 | 0 | 0.055 | 0 | 0.014 | 0 |
|  | id | 0.065 | 0 | 0.053 | 0 | 0.039 | 0 |
|  | ja | 0.107 | 0 | 0.063 | 0 | 0.061 | 0 |
|  | ko | 0.041 | 0 | 0.107 | 0 | 0.078 | 0 |
|  | ru | 0.044 | 0 | 0.071 | 0 | 0.012 | 0 |
| Linear predictors | ar | 0.09 | 0.32 | 0.057 | 0.15 | 0.042 | 0.46 |
|  | en | 0.041 | 0.36 | 0.027 | 0.17 | 0.034 | 0.59 |
|  | fi | 0.068 | 0.34 | 0.070 | -0.12 | 0.015 | 0.05 |
|  | id | 0.070 | 0.06 | 0.033 | 0.47 | 0.037 | 0.11 |
|  | ja | 0.108 | 0.10 | 0.09 | -0.36 | 0.045 | 0.26 |
|  | ko | 0.045 | 0.02 | 0.070 | 0.37 | 0.056 | 0.30 |
|  | ru | 0.047 | 0.10 | 0.049 | 0.33 | 0.013 | 0.08 |
| Gradient boosting | ar | 0.107 | 0.13 | 0.067 | 0.02 | 0.032 | 0.59 |
|  | en | 0.104 | 0.23 | 0.028 | 0.10 | 0.013 | 0.56 |
|  | fi | 0.070 | 0.32 | 0.069 | -0.11 | 0.009 | 0.33 |
|  | id | 0.090 | -0.25 | 0.046 | 0.26 | 0.022 | 0.47 |
|  | ja | 0.104 | 0.08 | 0.080 | -0.33 | 0.040 | 0.33 |
|  | ko | 0.046 | -0.04 | 0.082 | 0.26 | 0.077 | 0.06 |
|  | ru | 0.044 | 0.08 | 0.061 | 0.14 | 0.007 | 0.47 |
| Optimal Probe | ar | 0.103 (1) | 0.23 | 0.054 (3) | 0.10 | 0.037 (4) | 0.59 |
|  | en | 0.043 (1) | 0.35 | 0.025 (6) | 0.11 | 0.010 (3) | 0.63 |
|  | fi | 0.080 (12) | -0.01 | 0.039 (2) | 0.14 | 0.006 (1) | 0.59 |
|  | id | 0.046 (9) | 0.35 | 0.036 (3) | 0.21 | 0.025 (3) | 0.32 |
|  | ja | 0.007 (6) | 0.40 | 0.071 (7) | 0 | 0.023 (5) | 0.66 |
|  | ko | 0.034 (12) | 0 | 0.062 (8) | 0.13 | 0.073 (9) | 0.18 |
|  | ru | 0.041 (9) | 0.18 | 0.037 (7) | 0.14 | 0.004 (5) | 0.68 |
| Weakest Probe | ar | 0.140 (7) | 0 | 0.065 (5) | -0.02 | 0.079 (12) | 0 |
|  | en | 0.076 (8) | 0.04 | 0.030 (12) | 0.02 | 0.023 (12) | 0.17 |
|  | fi | 0.112 (5) | -0.13 | 0.051 (12) | -0.03 | 0.016 (12) | -0.03 |
|  | id | 0.062 (12) | 0.04 | 0.046 (12) | 0.04 | 0.034 (7) | 0.08 |
|  | ja | 0.094 (12) | 0.06 | 0.093 (10) | -0.14 | 0.063 (12) | 0.03 |
|  | ko | 0.036 (11) | -0.04 | 0.074 (6) | -0.07 | 0.104 (3) | 0.12 |
|  | ru | 0.057 (8) | -0.08 | 0.051 (12) | -0.07 | 0.016 (12) | -0.09 |
| Fine-tuned Glot500 | ar | 0.030 | - | 0.055 | - | 0.056 | - |
|  | en | 0.039 | - | 0.023 | - | 0.020 | - |
|  | fi | 0.079 | - | 0.044 | - | 0.011 | - |
|  | id | 0.045 | - | 0.038 | - | 0.027 | - |
|  | ja | 0.078 | - | 0.071 | - | 0.035 | - |
|  | ko | 0.034 | - | 0.074 | - | 0.070 | - |
|  | ru | 0.038 | - | 0.043 | - | 0.006 | - |

Table 6: Performance metrics for linguistic complexity sub-metric regression tasks across seven languages (Part 2: Verbal Edges, Lexical Density, N Tokens).

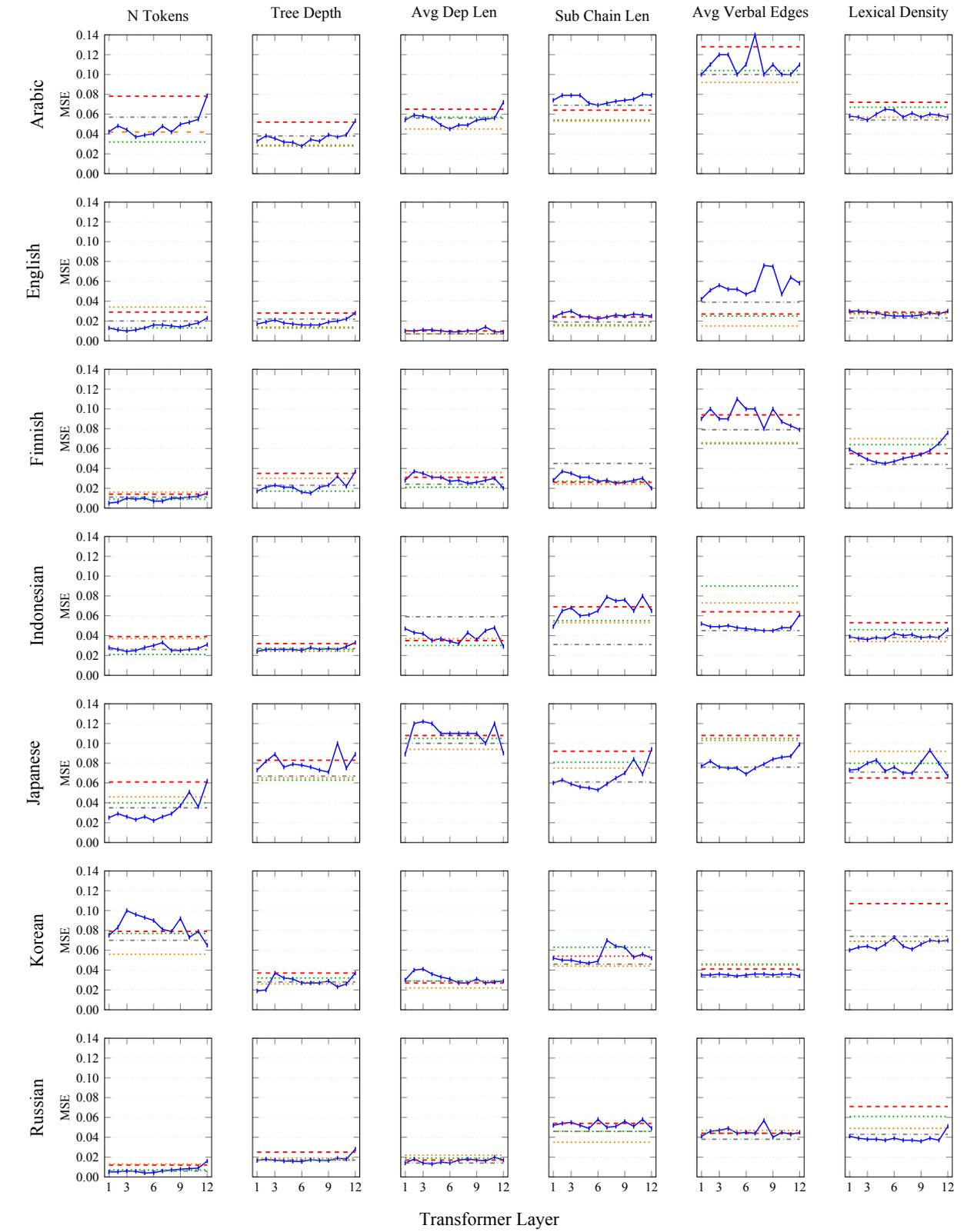

Figure 5: Performance metrics across languages and transformer layers